\documentclass[conference]{IEEEtran}
\IEEEoverridecommandlockouts
\usepackage{cite}
\usepackage{amsmath,amssymb,amsfonts}
\usepackage{algorithmic}
\usepackage{graphicx}
\usepackage{textcomp}
\usepackage{textgreek}
\usepackage{xcolor}
\usepackage{multirow}
\usepackage{booktabs} 
\usepackage{array}    
\usepackage[caption=false]{subfig} 
\usepackage{graphicx}             

\newcolumntype{C}{>{\centering\arraybackslash}m{0.105\textwidth}}

\def\BibTeX{{\rm B\kern-.05em{\sc i\kern-.025em b}\kern-.08em
    T\kern-.1667em\lower.7ex\hbox{E}\kern-.125emX}}
\begin{document}

\title{RetinaGuard: Obfuscating Retinal Age in Fundus Images for Biometric Privacy Preserving\\
\thanks{* Corresponding Author}
}

\author{\IEEEauthorblockN{1\textsuperscript{st} Zhengquan Luo}
\IEEEauthorblockA{\textit{Faculty of Data Science} \\
\textit{City University of Macau}\\
Macao, China }
\and
\IEEEauthorblockN{2\textsuperscript{nd} Chi Liu*}
\IEEEauthorblockA{\textit{Faculty of Data Science} \\
\textit{City University of Macau}\\
Macao, China }
\and
\IEEEauthorblockN{3\textsuperscript{rd} Dongfu Xiao}
\IEEEauthorblockA{\textit{Faculty of Data Science} \\
\textit{City University of Macau}\\
Macao, China }
\and
\IEEEauthorblockN{4\textsuperscript{th} Zhen Yu}
\IEEEauthorblockA{\textit{Faculty of lnformation Technology} \\
\textit{Monash University}\\
Melbourne, Victoria, Australia }
\and
\IEEEauthorblockN{5\textsuperscript{th} Yueye Wang}
\IEEEauthorblockA{\textit{Faculty of Health and Social Sciences} \\
\textit{Hong Kong Polytechnic University}\\
Hong Kong, China }
\and
\IEEEauthorblockN{6\textsuperscript{th} Tianqing Zhu}
\IEEEauthorblockA{\textit{Faculty of Data Science} \\
\textit{City University of Macau}\\
Macao, China}
}

\maketitle

\begin{abstract}
The integration of AI with medical images enables the extraction of implicit image-derived biomarkers for a precise health assessment. Recently, retinal age, a biomarker predicted from fundus images, is a proven predictor of systemic disease risks, behavioral patterns, aging trajectory and even mortality. However, the capability to infer such sensitive biometric data raises significant privacy risks, where unauthorized use of fundus images could lead to bioinformation leakage, breaching individual privacy. In response, we formulate a new research problem of biometric privacy associated with medical images and propose RetinaGuard, a novel privacy-enhancing framework that employs a feature-level generative adversarial masking mechanism to obscure retinal age while preserving image visual quality and disease diagnostic utility. The framework further utilizes a novel multiple-to-one knowledge distillation strategy incorporating a retinal foundation model and diverse surrogate age encoders to enable a universal defense against black-box age prediction models. Comprehensive evaluations confirm that RetinaGuard successfully obfuscates retinal age prediction with minimal impact on image quality and pathological feature representation. RetinaGuard is also flexible for extension to other medical image-derived biomarkers. 
\end{abstract}

\section{Introduction}
The integration of AI with medical images has become an accepted approach for computer-aided disease diagnosis and health assessment in clinical practice. While most methods use AI to detect pathological features, a growing trend focuses on predicting implicit biomarkers from images—such as biological age, sex, BMI, and smoking status\cite{wang2022predicting,adleberg2022predicting,duffy2022confounders}. These biomarkers, particularly biological age, can reflect an individual's health status and be exploited to profile one's behavioral tendencies and life habits \cite{jylhava2017biological,rutledge2022measuring,baecker2021machine,liu2022privacy}. For example, the biological age predicted from retinal fundus images, also known as \emph{retinal age}, has been proven an effective predictor of systemic disease risks (such as stroke and cardiovascular diseases), aging trajectory, and even mortality \cite{wang2024towards,zhu2023retinal,yu2023retinal,liu2019biological}.

The exploration of image-derived biomarkers significantly improves precise and personalized health management. However, the risks to individual biometric privacy arising from the predictions of these biomarkers are often underestimated \cite{QuestionTrustAI2024}. Established privacy regulations, such as the General Data Protection Regulation (GDPR), assert that any individual biometric data that can be inferred by AI models must be classified as highly sensitive information \cite{voigt2017eu}. Taking fundus images as an example, an unscrupulous insurance service provider may collect fundus images from users to ascertain indemnity, yet misuse them to assess users' underlying health conditions without their consent, breaching their privacy.

To safeguard against unauthorized access to such biometrics, it is imperative to implement privacy-enhancing techniques (PETs) for medical images prior to their dissemination to untrusted entities \cite{QuestionTrustAI2024}. Unfortunately, current research on this topic is notably limited. In addition to a general lack of privacy awareness regarding such implicit information, several technical challenges remain to impede progress. First, the biometric features are implicit and difficult to localize \cite{liu2019biological}; thus, conventional image-processing PETs such as blurring and mosaic are impractical while more effective biometric feature disentanglement and obfuscation are required. Second, PETs designed for biometric privacy must be universal against unseen biometric prediction models, assuming that most prediction models are in a black box and inaccessible to PET developers \cite{zhou2022adversarial}. Last, the privacy-utility trade-off in medical images is particularly challenging, as their diagnostic usage requires not only human-interpretable visual quality but also accurate representation of pathological features for AI analysis; thus, the commonly used PET of adversarial perturbation is less effective, as it inadvertently undermines disease predictability while perturbing biometric predictions.

To address this gap, this paper conceptualizes biometric privacy in medical imaging as a new problem, with mathematical formulations of the above challenges. For our pilot study, we focus on retinal age derived from color fundus images, as the rich RGB pixel information in these images further complicates this problem compared to grayscale modalities such as X-ray and MRI. We propose a novel privacy-enhancing framework, called \emph{RetinaGuard}, to mitigate the identified challenges. RetinaGuard employs a feature-level adjustable generative adversarial masking mechanism to obscure retinal age. To enhance the efficacy of this obfuscation against arbitrary black-box age models, we devise a multiple-to-one knowledge distillation strategy to learn shared age features from diverse surrogate age encoders for adversarial mask generation. We also adapt a retinal foundation model as the backbone of the student age encoder, enhancing both distillation efficiency and effectiveness with pre-trained general retinal features. Alongside optimizing mask generation and knowledge distillation, we impose a joint constraint on image quality and the integrity of pathological features to maintain image utility. Comprehensive evaluations across various fundus image datasets and retinal age models validate the protective capability of RetinaGuard, along with extensive analyses of the privacy-utility trade-off. Notably, RetinaGuard also offers high flexibility for extension to other types of image-associated biomarkers beyond retinal age.

\section{Related Work}
Privacy-enhancing technologies (PETs) in the medical AI domain have primarily focused on holistic image-level privacy preservation, such as digital masking \cite{yang2022digital} and federated learning \cite{shen2024federated}. However, these approaches treat the entire image as sensitive and operate at the pixel level, making them inadequate for feature-level biometric privacy concerns like retinal age prediction. Traditional image-level perturbation methods, while computationally straightforward, severely degrade visual fidelity and semantic integrity. For instance, Gaussian noise or blurring indiscriminately corrupts diagnostic information essential for clinical interpretation.

To mitigate effects on image semantics while preserving privacy, some approaches decompose images into RGB channels and compute average pixel intensities per channel, then randomize a subset of pixel values toward these averages \cite{wu2023identification}. However, such techniques remain limited to the pixel domain and cannot effectively balance privacy and diagnostic utility.

More sophisticated methods target privacy at the feature level rather than the pixel level. These recognize that sensitive biometric information is encoded in high-level semantic features, not raw pixels. Recent works propose feature-level privacy mechanisms for facial recognition and medical imaging. For example, k-anonymity mixers applied to fully connected layer features enable reconstruction while achieving de-identification through feature blending \cite{heinrich2024implicit}. Another approach computes contribution weights of latent codes to adjust noise injection, enabling more precise de-identification \cite{zhao2023unobtrusive}.

While these feature-level techniques achieve better privacy-utility trade-offs than pixel-level methods, they inadvertently compromise task-relevant features for downstream clinical applications. Crucially, such perturbations may introduce artifacts in critical regions—such as pathological lesions and vascular structures—undermining the reliability of diagnosis and segmentation. The indiscriminate nature of noise injection fails to distinguish between privacy-sensitive and diagnostically relevant features.

Building on these limitations, our work introduces an adversarial loss-guided noise generation framework that targets biometric privacy while preserving pathological feature integrity. By incorporating a dual optimization objective—maximizing biometric obfuscation and preserving disease-relevant representations—our approach achieves a more nuanced balance between privacy and utility, a critical requirement for clinical deployment.

\section{The RetinaGuard framework}
\subsection{The definition of biometric privacy in medical images}
We define the biometric privacy in medical images as the implicit biomarkers or sensitive attributes that can be predicted from medical images via AI models, e.g., $\mathcal{A}(\mathbf{X}):x\rightarrow a$, where $\mathcal{A}(\cdot)$ is a machine learning model, $x\in\mathbf{X}$ is a medical image and $a$ is a target biomarker. The goal of privacy protection is to manipulate the image, such that $\mathcal{A}$ incorrectly predicts the manipulated image. Particularly, in the context of medical imaging, the visual quality and the pathological features of the manipulated image must remain consistent with those of the raw image to ensure diagnostic efficacy for both human experts and AI disease models. This can be formulated as an optimization problem with the following objective: 

\begin{equation}
\begin{split}
    & \mathop{\arg\max}_{\mathcal{M}:x \rightarrow \hat{x}} ||\mathcal{A}(\hat{x})-\mathcal{A}(x)||,\ \\ & \quad s.t. \quad ||\hat{x}-x|| \leq \epsilon, \quad ||\mathcal{D}(\hat{x})-\mathcal{D}(x)|| \leq \tau,
    \label{eq1}
\end{split}
\end{equation}
where $\mathcal{M}(\cdot)$ is the manipulation function, $\mathcal{D}(\cdot)$ is a pathological feature encoder, and $\epsilon$ and $\tau$ are small scalars. We assume that $\mathcal{A}$ is completely inaccessible during optimization and expect that the optimal $\mathcal{M}(\cdot)$ can universally defend against arbitrary black-box biomarker models. This is a practical assumption since the privacy protector usually knows little about the surreptitious biomarker snooper.

\subsection{RetinaGuard}
\begin{figure*}[t]
    \centering
    \includegraphics[width=0.9\textwidth]{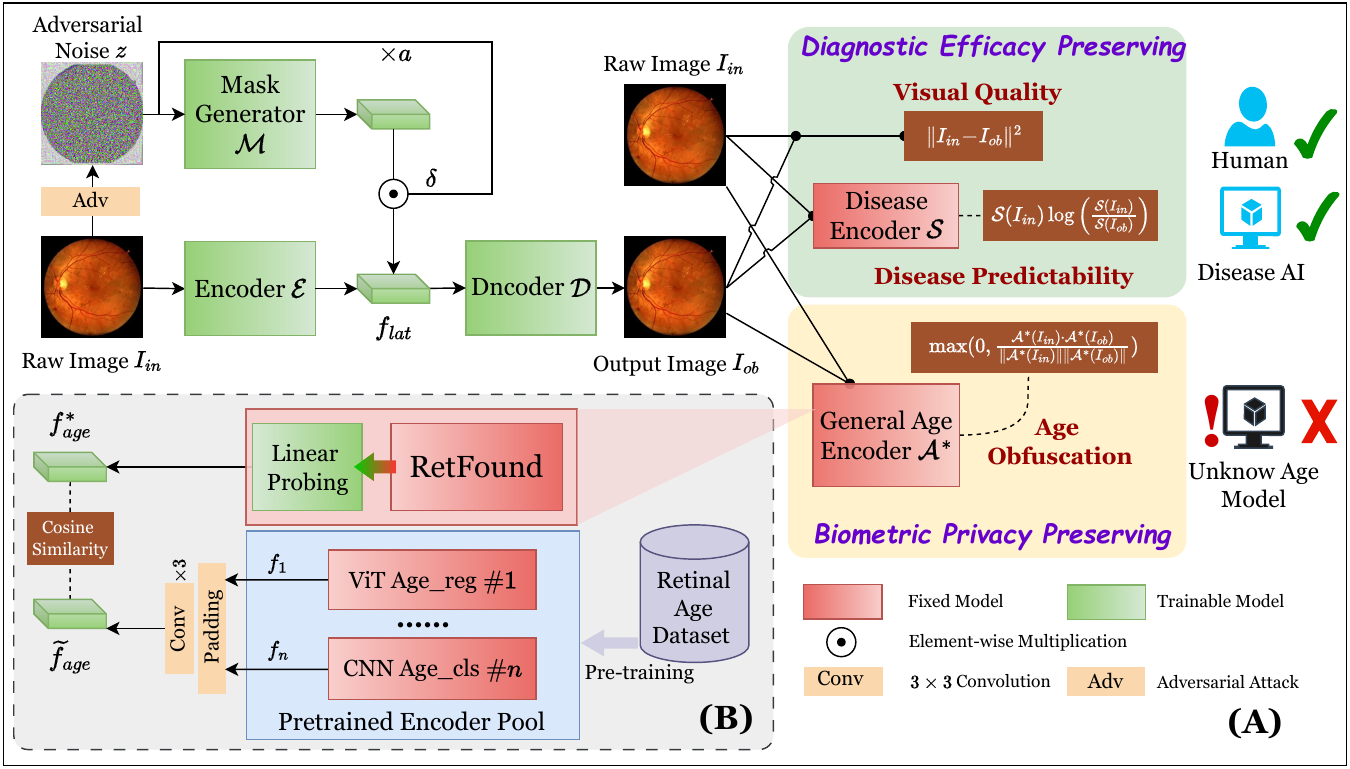}
    \caption{(A) The overview of RetinaGuard, which balances the trade-off between age privacy and diagnostic efficacy. (B) The knowledge distillation strategy for a general surrogate foundation age encoder.}
    \label{fig:framework}
\end{figure*}

We present RetinaGuard, a privacy-enhancing framework that tackles the biometric privacy problem in the context of retinal age. It consists of two key components: 1) a feature-level adjustable generative adversarial masking mechanism that obscures retinal age while preserving image quality and the integrity of disease features; and 2) a complementary knowledge distillation strategy that creates a shared age feature space with low training overhead, which helps enhance the transferability of the age mask across various black-box age prediction models without accessing their weights or gradients. Fig.\ref{fig:framework}-(A) illustrates the overview of RetinaGuard. 

\noindent \textbf{Feature-level Adversarial Masking}\quad The obfuscation of retinal age relies on operating latent features rather than raw images, as it is easier and more accurate to extract implicit bioinformation from latent features than from pixel domain. Thus, a deep encoder is employed to transform the input image to a latent feature, denoted as $f_{lat}=\mathcal{E}(I_{in})$. In addition to the image branch, we employ a trainable adversarial mask generator $\mathcal{M}(\cdot)$ to create a feature-level adversarial mask from a noise input $z$, which is then applied to the latent feature via element-wise multiplication. To accelerate the search for the optimal mask, we feed $\mathcal{M}(\cdot)$ a specific adversarial noise crafted from the input image instead of random noise. The entire masking process is denoted as $f_{ob}=f_{lat} \odot \mathcal{M}\left(z(I_{in})\right)$, where $f_{ob}$ is the obfuscated latent feature. Then, a decoder reconstructs the age-obfuscated image from $f_{ob}$, denoted as $I_{ob} = \mathcal{D}(f_{ob})$. For inference, a scaler $\delta=a \cdot \hat{z}$ is applied to $\mathcal{M}$ as $\delta \odot \mathcal{M}$ to adjust the privacy-utility trade-off as per user preferences, where $a\in(0,1]$ is a user-defined factor and $\hat{z}$ is the z-score normalization of the input adversarial noise $z$. 

To satisfy the privacy protection objectives and constraints in Eq. \ref{eq1}, we devise the following joint loss function for training $\{\mathcal{E}, \mathcal{M}, \mathcal{D}\}$.
\begin{equation}
\begin{split}
    L_{pp} = &\lambda \underbrace{\max(0, \vphantom{\frac{\mathcal{A}^*(I_{in}) \cdot \mathcal{A}(I_{ob})}{\|\mathcal{A}^*(I_{in})\| \|\mathcal{A}(I_{ob}\|}}\frac{\mathcal{A}^*(I_{in}) \cdot \mathcal{A}^*(I_{ob})}{\|\mathcal{A}^*(I_{in})\| \|\mathcal{A}^*(I_{ob})\|})}_{\text{Privacy preserving}} + \\ &\phi \underbrace{ \vphantom{\frac{\mathcal{A}^*(I_{in}) \cdot \mathcal{A}(I_{ob})}{\|\mathcal{A}^*(I_{in})\| \|\mathcal{A}(I_{ob}\|}} \|I_{in}\!-\!I_{ob}\|^{2} }_{\text{Visual quality}} + \\ &\underbrace{(1\!-\!\lambda\!-\!\phi)  \vphantom{\frac{\mathcal{A}^*(I_{in}) \cdot \mathcal{A}(I_{ob})}{\|\mathcal{A}^*(I_{in})\| \|\mathcal{A}(I_{ob}\|}} \mathcal{S}(I_{in}) \log\left(\frac{\mathcal{S}(I_{in})}{\mathcal{S}(I_{ob})}\right) }_{\text{Disease feature integrity}},
\end{split}
\label{eq2}
\end{equation}
where the first term defines the cosine similarity between age features encoded from $I_{in}$ and $I_{ob}$ using a fixed age encoder $\mathcal{A}^*$; the second calculates the mean square error between $I_{in}$ and $I_{ob}$; the last measures the KL-divergence between pathological features of the two images encoded by a fixed disease encoder $\mathcal{S}$; $\lambda$ and $\phi$ are balance factors. The loss function aims to minimize age feature similarity between $I_{in}$ and $I_{ob}$ while maximizing their pixel-level and disease feature consistencies. By optimizing this loss, we can achieve age obfuscation while maintaining image visual quality and diagnostic utility. Note that the selection of $\mathcal{S}$ is flexible, depending on the pathological features the user wishes to preserve, while $\mathcal{A}^*$ is introduced next. 

\noindent \textbf{Knowledge Distillation of Shared Age Feature}\quad The above optimization requires a fixed pre-trained age encoder $\mathcal{A}^*$, while not accessing any target age models under the black-box assumption. Furthermore, the encoder must be sufficiently general to create a shared age feature space, thereby making the adversarial mask a universal defense against arbitrary unknown age prediction models. We employ a surrogate age encoder to address the black-box problem, inspired by black-box adversarial attacks that employ surrogate models to mimic the behavior of targets \cite{papernot2017practical}. However, relying solely on one specific surrogate encoder can introduce model bias and limit generalization, while utilizing multiple encoders significantly complicates the optimization of Eq \ref{eq2}. To tackle this, we propose a novel knowledge distillation strategy to learn a general surrogate foundation age encoder from diverse pre-trained age encoders with different architectures (i.e., CNNs and ViTs) and prediction tasks (i.e., age regression and classification).

As shown in Fig.\ref{fig:framework}-(B), we utilize the fixed encoder of the pre-trained retinal foundation model RETFound \cite{zhou2023foundation} as the backbone of the surrogate encoder $\mathcal{A}^*$ and add a trainable linear probing layer on top for fine-tuning. This design, leveraging RETFound's power in extracting general retinal features from fundus images, enables more efficient and focused training for retinal age knowledge distillation compared to training a small encoder from scratch. 

The distillation is also performed at the feature level using contrastive learning. Let $\{\widetilde{\mathcal{A}}_i\}$ be a pool of pre-trained age encoders to be distilled, we extract the age feature from each encoder, and align their dimensions using zero-padding. The dimension-aligned features are then stacked and feed to three convolutional layers for integration, resulting in a fused feature $\widetilde{f}_{age}$. Then we optimize the cosine similarity between $\widetilde{f}_{age}$ and $f^*_{age}$, denoted as
\begin{equation}
    L_{dil}=1-\frac{\widetilde{f}_{age} \cdot f^*_{age}}{\|\widetilde{f}_{age}\| \|f^*_{age}\|},
\end{equation}
where $f^*_{age}$ is the feature output by $\mathcal{A}^*$. 

\section{Experiments}
\subsection{Settings}
\noindent \textbf{Datasets}\quad Four fundus image datasets are employed for training: ODIR\footnote{https://www.kaggle.com/datasets/andrewmvd/ocular-disease-recognition-odir5k}, UKB, Kaggle-DR\footnote{https://www.kaggle.com/competitions/diabetic-retinopathy-detection}, and Standardized Multi-Channel Dataset for Glaucoma (SMDG-19)\footnote{https://www.kaggle.com/datasets/deathtrooper/multichannel-glaucoma-benchmark-dataset}. ODIR contains 8,000 images with age data (ranging from 1 to 91, mean 57.85) and multiple disease labels. We use the officially partitioned training set to train the RetinaGuard model. UKB is used for training surrogate retinal age encoders \(\{\widetilde{\mathcal{A}}_i\}\). It consists of 10,000 images randomly selected from healthy individuals aged 40 to 69 in the UK Biobank cohort \cite{bycroft2018uk}, as biological age is typically assessed in healthy populations. For the pathological feature, we train the disease encoder \(\mathcal{S}\) using the Kaggle-DR and SMDG datasets for DR and GON, respectively. Kaggle-DR contains 35,126 retinal images annotated for diabetic retinopathy (DR), while SMDG comprises 12,449 retinal images aggregated from 19 publicly available, standardized glaucoma fundus image datasets, including 6,392 full-color annotated fundus images. For the test set, we use the official ODIR test set and an independent set from the RAE dataset \cite{yu2023retinal}, which combines images from UK Biobank and three Chinese cohorts. Inclusion criteria are applied based on image quality and prediction errors of the raw images to avoid evaluation bias from factors like poor image quality. Dataset details are shown in Table \ref{tab:dataset}.

\begin{table*}[htbp]
  \centering
  \caption{The test sets and inclusion criteria for evaluation.}
  \label{tab:dataset}%
  \resizebox{1\textwidth}{!}{
        \setlength{\tabcolsep}{8mm}{
    \begin{tabular}{clll}
      \toprule
      Source & Usage & N & Inclusion criteria \\
      \midrule
      ODIR & Age prediction & 169 & Raw image age prediction MAE $\leq$ 3.0 \\
      & Quality assessment & 812 & Raw image with good quality; DR prediction ACC $\geq$ 0.85 \\
      & Disease prediction & 812 & Same as above \\
      \midrule
      RAE & Age prediction & 150 & Raw image with good quality; Age prediction MAE $\leq$ 3.0 \\
      & Quality assessment & 150 & Same as above \\
      \bottomrule
    \end{tabular}}}%
\end{table*}

\noindent \textbf{Setup}\quad We employ ResUnet \cite{zhang2018road} as the encoder-decoder in RetinaGuard, and ResNet18 as the disease encoder $\mathcal{S}(\cdot)$. The adversarial noise $z$ is crafted from an ImageNet-pre-trained ResNet18 using Projected Gradient Descent (PGD) \cite{madry2018towards}. The mask generator $\mathcal{M}(\cdot)$ consists of three $3 \times 3$ convolution layers. The default values of $a$, $\lambda$, and $\phi$ are 0, 0.4, and 0.4 for training. Images are resized to $256\times256$ for CNN age encoders and $384\times384$ for ViT age encoders. Random cropping and rotation are applied with probability 0.2 during training.

\noindent \textbf{Metrics}\quad We assess age prediction using mean absolute error (MAE) and R-squared score (R2). MAE measures the average prediction error of an age model, while R2 characterizes the correlation between ground-truth and predicted ages. We evaluate disease prediction and image quality using DR classification accuracy (DR-ACC), GON classification accuracy (GON-ACC), structural similarity index measure (SSIM), and fundus vessel segmentation performance measured by Intersection over Union (IoU) \cite{liu2022full}.

\noindent \textbf{Pre-trained age models and baselines}\quad We employ four public age models as targets: two classification and two regression models based on ResNet-50 and ViT-B-16, denoted as ResNet-cls, ResNet-reg, ViT-cls, and ViT-reg. All models were pre-trained on the RAE dataset by Yu et al. \cite{yu2023retinal}. Our privacy-preserving baselines include a Gaussian blur model (BLUR) with kernel size $(7,7)$, a Gaussian noise model (NOISE) with mean 0 and variance 8, a PGD adversarial attack model (ADV) optimized using the target ResNet-cls age model with perturbation magnitude $8/255$, a DP-BLUR model \cite{fan2019differential}, a de-identification model \cite{heinrich2024implicit}, and a K-Anonymity model \cite{zhao2023unobtrusive}.

\begin{figure}[htbp]
    \centering
    \includegraphics[width=\linewidth]{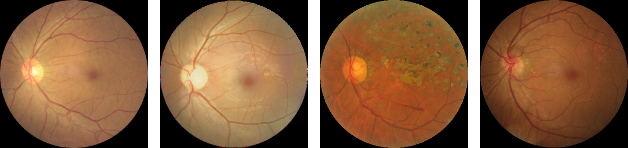}
    \caption{Synthetic fundus images generated by RetinaGuard. (a) Normal fundus without pathology; (b)-(d) Pathological cases.}
    \label{fig:synthetic_examples}
\end{figure}

\begin{table}[htbp]
    \centering
    \caption{Quantitative evaluation of synthetic fundus image quality}
    \label{tab:quality_metrics}
    \begin{tabular}{l c c c c c}
        \toprule
        \textbf{Metric} & \textbf{SSIM} & \textbf{ACC} & \textbf{Recall} & \textbf{Precision} & \textbf{F1-Score} \\
        \midrule
        Objective Quality & 0.948 & -- & -- & -- & -- \\
        Human Perception & -- & 0.542 & 0.535 & 0.495 & 0.511 \\
        \bottomrule
    \end{tabular}
\end{table}

\begin{table*}[htbp]
  \centering
  \caption{Retinal age prediction performance using four pre-trained models on ODIR and RAE datasets. Higher MAE and lower R\textsuperscript{2} indicate stronger privacy protection.}
  \label{tab:age}
  \resizebox{1\textwidth}{!}{
        \setlength{\tabcolsep}{5mm}{
  \begin{tabular}{@{}l l *{8}{c}@{}}
    \toprule
    \textbf{Dataset} & \textbf{Method} & 
    \multicolumn{2}{c}{\textbf{ResNet-cls}} & 
    \multicolumn{2}{c}{\textbf{ResNet-reg}} & 
    \multicolumn{2}{c}{\textbf{ViT-cls}} & 
    \multicolumn{2}{c}{\textbf{ViT-reg}} \\
    \cmidrule(lr){3-4} \cmidrule(lr){5-6} \cmidrule(lr){7-8} \cmidrule(lr){9-10}
    & & \textbf{MAE$\uparrow$} & \textbf{R\textsuperscript{2}$\downarrow$} & 
        \textbf{MAE$\uparrow$} & \textbf{R\textsuperscript{2}$\downarrow$} & 
        \textbf{MAE$\uparrow$} & \textbf{R\textsuperscript{2}$\downarrow$} & 
        \textbf{MAE$\uparrow$} & \textbf{R\textsuperscript{2}$\downarrow$} \\
    \midrule
    ODIR & RAW             & 2.45  & 0.70  & 2.99  & 0.72  & 1.94  & 0.83  & 1.69  & 0.91  \\
     & BLUR            & 3.13  & 0.57  & 4.18  & 0.48  & 4.30  & 0.38  & 2.51  & 0.79  \\
     & DP-BLUR         & 1.81  & 0.84  & 2.14  & 0.83  & 2.09  & 0.83  & 2.83  & 0.74  \\
     & NOISE           & 2.61  & 0.67  & 3.46  & 0.64  & 2.01  & 0.84  & 1.78  & 0.90  \\
     & ADV             & 3.12  & \textbf{0.46} & 4.01  & 0.51  & 2.03  & 0.83  & 1.88  & 0.87  \\
     & K-ANONYMITY     & 2.54  & 0.60  & 2.42  & 0.78  & 3.47  & 0.53  & 2.75  & 0.74  \\
     & DE-IDENTIFICATION & 3.16  & 0.49  & 2.96  & 0.71  & 2.02  & 0.82  & 2.19  & 0.81  \\
     & RetinaGuard     & \textbf{3.52} & 0.51  & \textbf{4.48} & \textbf{0.37} & \textbf{7.62} & \textbf{-0.69} & \textbf{4.43} & \textbf{0.32} \\
    \midrule
    RAE  & RAW             & 2.06  & 0.92  & 1.91  & 0.94  & 1.65  & 0.95  & 0.01  & 1.00  \\
      & BLUR            & 2.43  & 0.88  & 1.95  & 0.94  & 2.04  & 0.93  & 0.83  & 0.99  \\
      & DP-BLUR         & 4.63  & 0.62  & 5.79  & 0.58  & 1.83  & 0.95  & 1.25  & 0.97  \\
      & NOISE           & 2.44  & 0.88  & 1.93  & 0.94  & 1.69  & 0.95  & 0.21  & 1.00  \\
      & ADV             & 2.48  & 0.87  & 2.49  & 0.90  & 1.83  & 0.94  & 0.69  & 0.99  \\
      & K-ANONYMITY     & 5.95  & 0.26  & 3.83  & 0.79  & 3.59  & 0.79  & 2.73  & 0.90  \\
      & DE-IDENTIFICATION & 9.16  & -0.50 & 5.92  & 0.53  & 1.74  & 0.96  & 1.75  & 0.95  \\
      & RetinaGuard     & \textbf{9.20} & \textbf{-0.54} & \textbf{6.62} & \textbf{0.40} & \textbf{6.10} & \textbf{0.50} & \textbf{3.94} & \textbf{0.76} \\
    \bottomrule
  \end{tabular}}}
\end{table*}

\subsection{Results}

\noindent \textbf{Visual and Perceptual Quality Assessment}\quad Figure \ref{fig:synthetic_examples} and Table \ref{tab:quality_metrics} present a comprehensive evaluation of the visual fidelity and perceptual realism of our synthesized fundus images. We randomly selected 20 real and 20 generated images from ODIR and RAE datasets for comparative analysis.

For Objective Quality Metrics, The synthesized images achieve a high SSIM score of 0.948 (Table \ref{tab:quality_metrics}), indicating excellent structural similarity to real fundus images. This result aligns with the visually imperceptible differences observed in Figure \ref{fig:synthetic_examples}, where both normal retinas (first column) and pathological cases (second to fourth columns) exhibit realistic vascular patterns, macular structures, and pathognomonic lesions such as hard exudates.

For Subjective Human Perception, three ophthalmologists performed a blind classification task on 40 images (20 real, 20 synthetic). Their average accuracy was 0.542 (chance = 0.5), with an F1 score of 0.511, precision of 0.495, and recall of 0.535—indicating performance barely above random guessing. Two doctors showed balanced discrimination (precision/recall: 0.48–0.57); the third tended to classify more images as synthetic (precision=0.42, recall=0.61). 

The near-chance-level discrimination performance (ACC=0.542) suggests that our method generates clinically plausible fundus images capable of mimicking subtle disease-specific features. This is particularly critical for privacy preservation, as it ensures that adversarial perturbations do not introduce artifacts that could expose synthetic origins. 

\noindent \textbf{Effect of Age Obfuscation}\quad Table \ref{tab:age} presents the age obfuscation results. RetinaGuard outperforms other baselines in both ODIR and RAE test sets, achieving the highest MAE scores and lowest R2 scores across different age models, except for the R2 score of ResNet-cls perturbed by ADV in ODIR. This occurs because the ADV model is a white-box perturbation that is optimized via directly descending the ResNet-cls gradients, whereas RetinaGuard is a completely black-box obfuscation without any access to pre-trained age models. In most cases, such as ResNet-cls in RAE and ViT-cls in both ODIR and RAE, the obfuscation effects of RetinaGuard greatly exceed baseline performance. Notably, ViT-cls in ODIR and ResNet-cls in RAE achieve negative R2 scores with RetinaGuard, indicating no statistical correlation exists between the ground-truth age labels and the predicted ages after RetinaGuard processing, meaning the retinal age information has been entirely removed from the processed image. 

The results, on the other hand, reveal the vulnerability of different model architectures in predicting implicit biomarkers, with ResNet demonstrating a generally better robustness against perturbations in retinal age prediction. For example, the average changes of the MAE and R2 scores over all perturbations are $+0.42$ and $-0.03$, slightly differing from the raw images. This could offer insights into future developments of robust image biomarker prediction models.

\noindent \textbf{Privacy-Utility Trade-off}\quad Table \ref{tab:tradeoff_transposed} illustrates the trade-offs between age privacy (AMAE) and diagnostic utility (SSIM, DR-ACC, GON-ACC, IoU) in terms of visual quality and disease feature integrity across different obfuscation models. Regarding image quality after obfuscation, RetinaGuard achieves suboptimal SSIM scores compared to all baselines in the RAE and ODIR dataset, but preserving vascular structure integrity achieves the higher IoU scores (0.874 for ODIR dataset 0.885 for RAE). However, while Blur increases the average retinal age MAE by only $+0.306$, RetinaGuard raises it by $+4.902$, indicating a more thorough age obfuscation. Overall, RetinaGuard effectively obfuscates age in fundus images while maintaining critical vascular structures, demonstrating a commendable balance between privacy and diagnostic utility. We also provide several samples in Figure \ref{fig:method_comparison} for visualization.

With regard to disease prediction, in the ODIR dataset, the DR-ACC, GON-ACC and IoU score for disease prediction is higher than all baselines, indicating that RetinaGuard will not compromise pathological feature representation or vascular structure integrity. Specifically, RetinaGuard achieves an IoU of 0.874, significantly outperforming Blur (0.443), ADV (0.027), and even K-Anonymity (0.867). Interestingly, RetinaGuard-processed images yield better DR-ACC scores than raw images. This improvement may be due to the disease encoder, pre-trained on the Kaggle-DR dataset, exhibiting inherent demographic bias when applied to different datasets. RetinaGuard addresses this by disentangling and suppressing age features, thereby refining disease features and enhancing the encoder's generalization ability. Similar observations regarding the impact of demographic bias on disease classifier generalizability  were reported by Yang et al. \cite{yang2024limits}, suggesting a potential avenue for future research.

\begin{table*}[htbp]
  \centering
  \caption{Image quality, disease feature integrity, and retinal age obfuscation performance on ODIR and RAE datasets. Higher values indicate better performance for all metrics ($\uparrow$). N/A: Not Applicable.}
  \label{tab:tradeoff_transposed}
  \resizebox{1\textwidth}{!}{
        \setlength{\tabcolsep}{4.5mm}{
  \begin{tabular}{@{}l *{9}{c}@{}}
    \toprule
    \textbf{Method} & 
    \multicolumn{5}{c}{\textbf{ODIR}} & 
    \multicolumn{4}{c}{\textbf{RAE}} \\
    \cmidrule(lr){2-6} \cmidrule(lr){7-10}
    & \textbf{SSIM$\uparrow$} & \textbf{DR-ACC$\uparrow$} & \textbf{GON-ACC$\uparrow$} & \textbf{AMAE$\uparrow$} & \textbf{IoU$\uparrow$} &
      \textbf{SSIM$\uparrow$} & \textbf{ACC$\uparrow$} & \textbf{AMAE$\uparrow$} & \textbf{IoU$\uparrow$} \\
    \midrule
    RAW(ref.)             & 1.000 & 0.906 & 0.911 & 2.267 & 1.000 & 1.000 & N/A   & 1.408 & 1.000 \\
    \midrule
    BLUR            & 0.942 & 0.908 & 0.886 & 3.531 & 0.443 & 0.933 & N/A   & 1.813 & 0.449 \\
    DP-BLUR         & 0.924 & 0.726 & 0.726 & 2.240 & 0.714 & 0.920 & N/A   & 2.583 & 0.727 \\
    NOISE           & 0.934 & 0.882 & 0.871 & 2.463 & 0.432 & 0.918 & N/A   & 1.569 & 0.581 \\
    ADV             & 0.905 & 0.875 & 0.852 & 2.760 & 0.027 & 0.908 & N/A   & 1.873 & 0.115 \\
    K-ANONYMITY     & \textbf{0.972} & 0.726 & 0.768 & 2.480 & 0.867 & \textbf{0.970} & N/A & 3.962 & 0.862 \\
    DE-IDENTIFICATION & 0.920 & 0.702 & 0.691 & 0.778 & 0.705 & 0.890 & N/A & 1.023 & 0.735 \\
    RetinalGuard    & 0.964 & \textbf{0.916} &  \textbf{0.905} & \textbf{4.902} & \textbf{0.874} & 0.929 & N/A & \textbf{5.900} & \textbf{0.885} \\
    \bottomrule
  \end{tabular}}}
\end{table*}

\begin{figure*}[!t]
  \centering
  \setlength{\tabcolsep}{2pt}        
  \renewcommand{\arraystretch}{1.05} 
  \begin{tabular}{@{}m{0.08\textwidth}@{} *{8}{C}@{}}
    & \rotatebox[origin=c]{45}{\scriptsize RAW}
    & \rotatebox[origin=c]{45}{\scriptsize Blur-based}
    & \rotatebox[origin=c]{45}{\scriptsize DP-Blur}
    & \rotatebox[origin=c]{45}{\scriptsize Noise}
    & \rotatebox[origin=c]{45}{\scriptsize Adversarial}
    & \rotatebox[origin=c]{45}{\scriptsize ReID}
    & \rotatebox[origin=c]{45}{\scriptsize K-Anon}
    & \rotatebox[origin=c]{45}{\scriptsize RetinaGuard} \\
    
    \centering ODIR \\ w/o DR &
    \includegraphics[width=\linewidth, height=2.0cm, keepaspectratio]{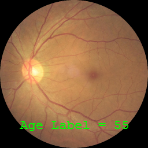} &
    \includegraphics[width=\linewidth, height=2.0cm, keepaspectratio]{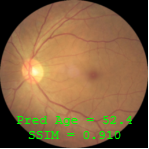} &
    \includegraphics[width=\linewidth, height=2.0cm, keepaspectratio]{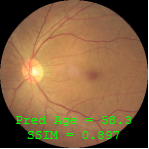} &
    \includegraphics[width=\linewidth, height=2.0cm, keepaspectratio]{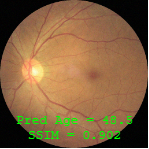} &
    \includegraphics[width=\linewidth, height=2.0cm, keepaspectratio]{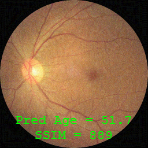} &
    \includegraphics[width=\linewidth, height=2.0cm, keepaspectratio]{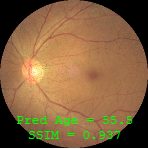} &
    \includegraphics[width=\linewidth, height=2.0cm, keepaspectratio]{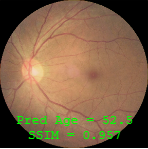} &
    \includegraphics[width=\linewidth, height=2.0cm, keepaspectratio]{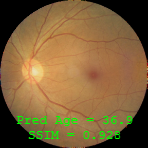} \\
    
    \centering ODIR \\ w/ DR &
    \includegraphics[width=\linewidth, height=2.0cm, keepaspectratio]{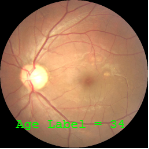} &
    \includegraphics[width=\linewidth, height=2.0cm, keepaspectratio]{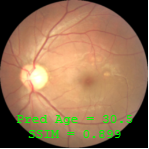} &
    \includegraphics[width=\linewidth, height=2.0cm, keepaspectratio]{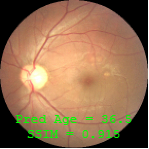} &
    \includegraphics[width=\linewidth, height=2.0cm, keepaspectratio]{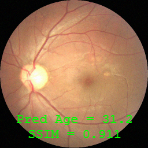} &
    \includegraphics[width=\linewidth, height=2.0cm, keepaspectratio]{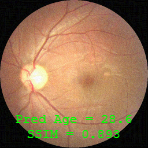} &
    \includegraphics[width=\linewidth, height=2.0cm, keepaspectratio]{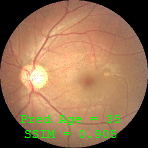} &
    \includegraphics[width=\linewidth, height=2.0cm, keepaspectratio]{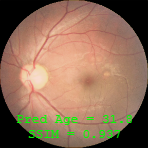} &
    \includegraphics[width=\linewidth, height=2.0cm, keepaspectratio]{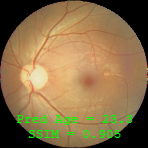} \\
    
    \centering RAE &
    \includegraphics[width=\linewidth, height=2.0cm, keepaspectratio]{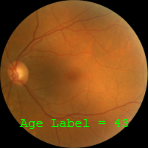} &
    \includegraphics[width=\linewidth, height=2.0cm, keepaspectratio]{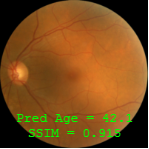} &
    \includegraphics[width=\linewidth, height=2.0cm, keepaspectratio]{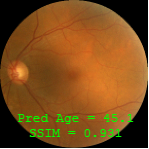} &
    \includegraphics[width=\linewidth, height=2.0cm, keepaspectratio]{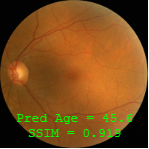} &
    \includegraphics[width=\linewidth, height=2.0cm, keepaspectratio]{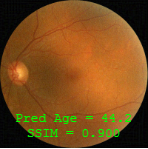} &
    \includegraphics[width=\linewidth, height=2.0cm, keepaspectratio]{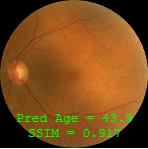} &
    \includegraphics[width=\linewidth, height=2.0cm, keepaspectratio]{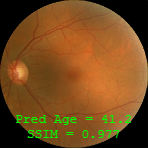} &
    \includegraphics[width=\linewidth, height=2.0cm, keepaspectratio]{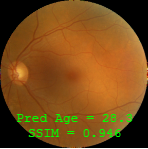} \\
  \end{tabular}
  
  \vspace{4pt}
  \captionsetup{font=small, labelfont=bf}
  \caption{Comparative analysis of image quality degradation across age obfuscation methods. Each column represents a different method: (a) Original image, (b) Blur-based method, (c) DP-Blur, (d) Noise injection, (e) Adversarial perturbation, (f) De-identification, (g) K-Anonymity, (h) RetinaGuard.}
  \label{fig:method_comparison}
\end{figure*}

\begin{table*}
  \centering
  \caption{Performance comparison of single-model, dual-model, and four-model (RetinalGuard) configurations for retinal age obfuscation. Higher values indicate better performance for all metrics ($\uparrow$). N/A: Not Applicable.}
  \label{tab:Configuration}
  \resizebox{1\textwidth}{!}{
        \setlength{\tabcolsep}{4.5mm}{
  \begin{tabular}{@{}l *{9}{c}@{}}
    \toprule
    \textbf{Configuration} & 
    \multicolumn{5}{c}{\textbf{ODIR}} & 
    \multicolumn{4}{c}{\textbf{RAE}} \\
    \cmidrule(lr){2-6} \cmidrule(lr){7-10}
    & \textbf{SSIM$\uparrow$} & \textbf{DR-ACC$\uparrow$} & \textbf{GON-ACC$\uparrow$} & \textbf{AMAE$\uparrow$} & \textbf{IoU$\uparrow$} &
      \textbf{SSIM$\uparrow$} & \textbf{ACC$\uparrow$} & \textbf{AMAE$\uparrow$} & \textbf{IoU$\uparrow$} \\
    \midrule
    single-model  & \textbf{0.950} & 0.915 & 0.895 & 4.610 & 0.857 & 0.943 & N/A & 5.499 & 0.860 \\
    dual-model    & \textbf{0.950} & 0.910 & 0.898 & 4.751 & 0.856 & \textbf{0.944} & N/A & 5.435 & 0.860 \\
    four-model (RetinalGuard) & \textbf{0.950} & \textbf{0.916} & \textbf{0.905} & \textbf{6.064} & \textbf{0.874} & \textbf{0.944} & N/A & \textbf{8.490} & \textbf{0.885} \\
    \bottomrule
  \end{tabular}}}
\end{table*}

\begin{figure*}
  \centering
  \subfloat[ODIR]{\includegraphics[width=0.35\textwidth]{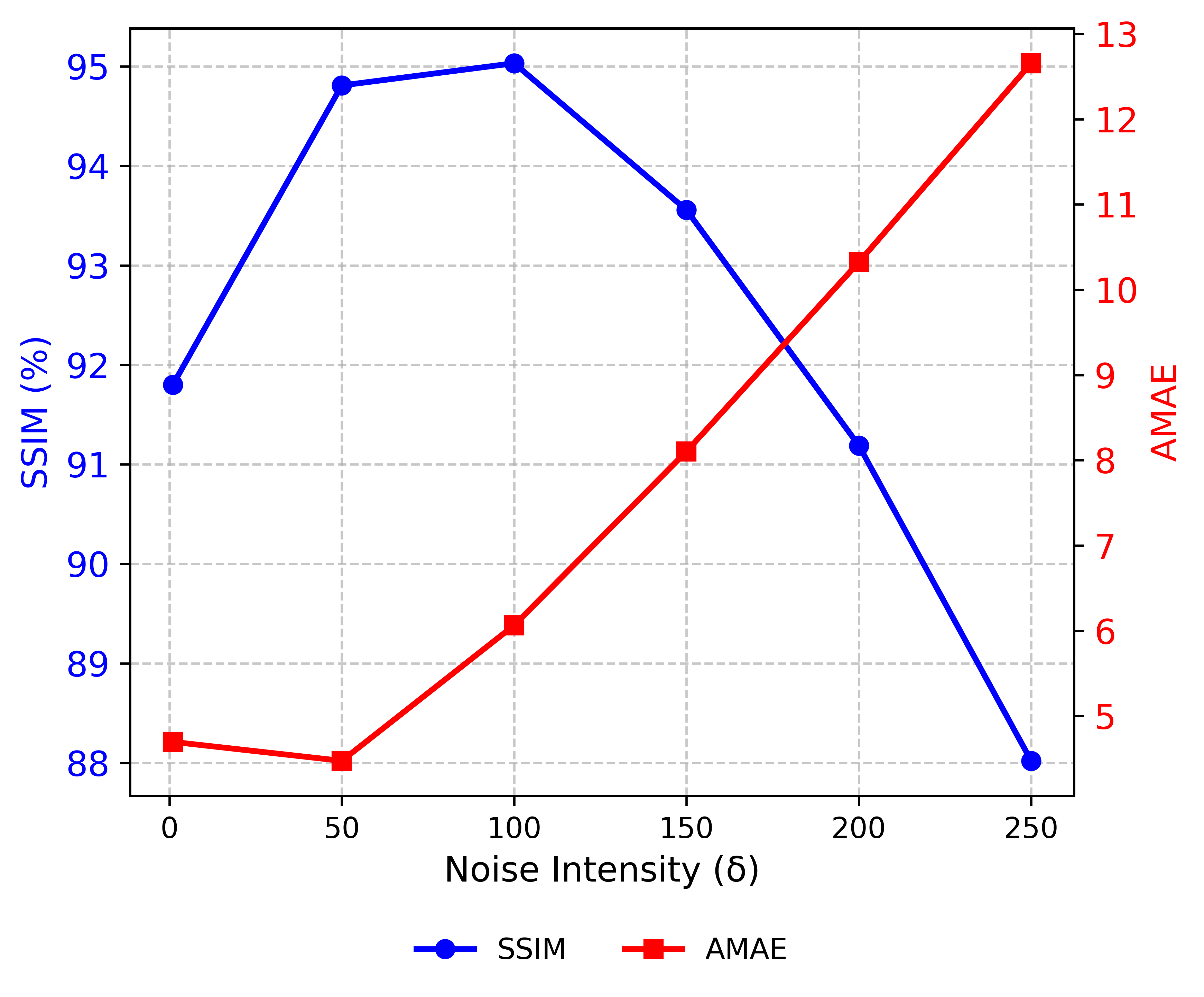}}\hfil
  \subfloat[RAE]{\includegraphics[width=0.35\textwidth]{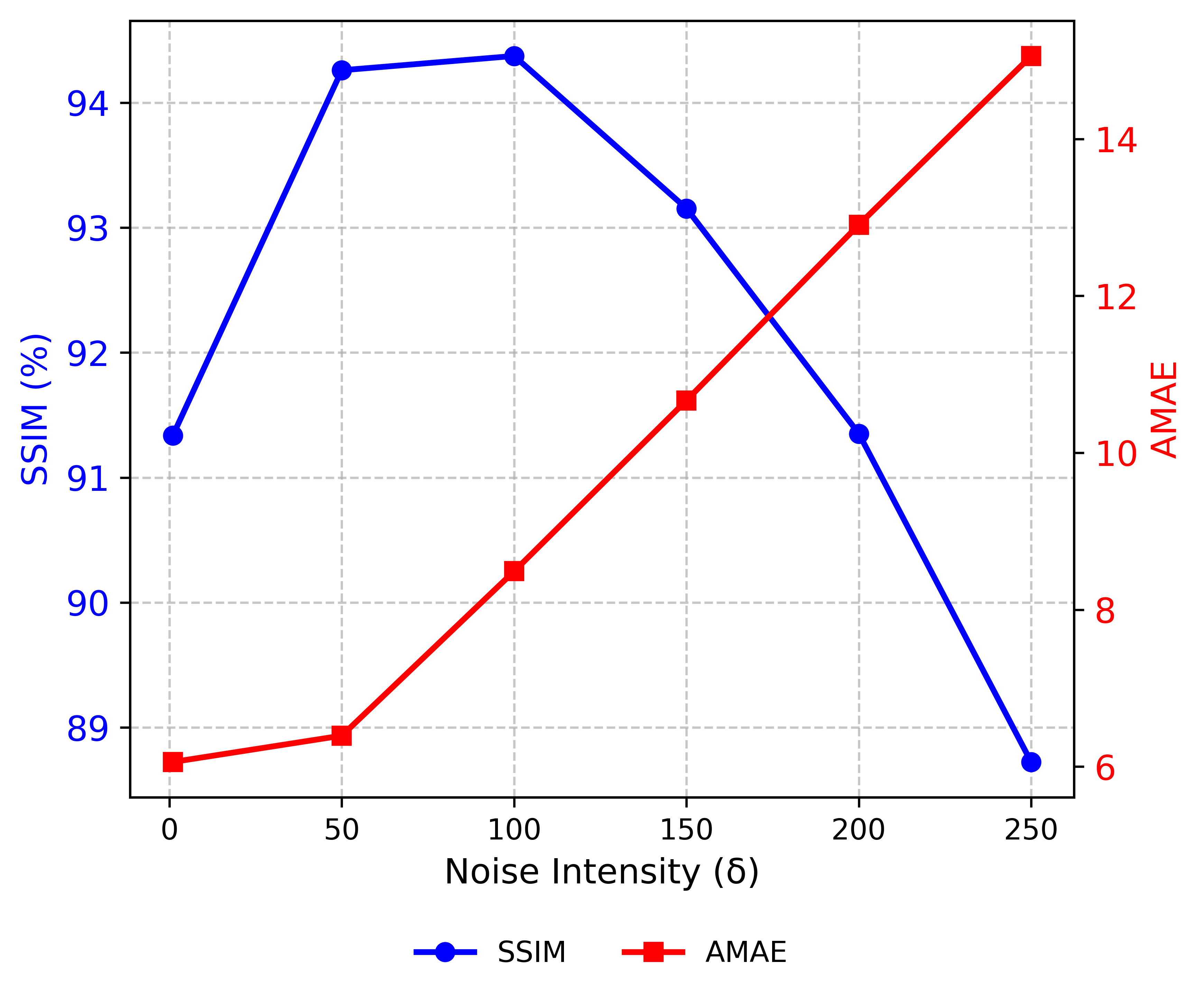}}
  \caption{Influence of noise intensity ($\delta$) on privacy preservation (AMAE) and diagnostic utility (SSIM) in the ODIR and RAE datasets.}
  \label{fig:detal}
\end{figure*}

\begin{figure*}
  \centering
  \subfloat[Reference]{\includegraphics[width=0.135\textwidth]{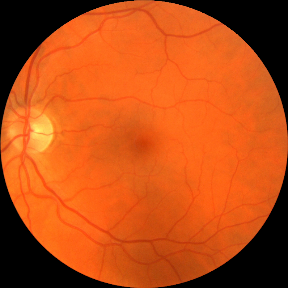}}\hfil
  \subfloat[$\delta=1$]{\includegraphics[width=0.135\textwidth]{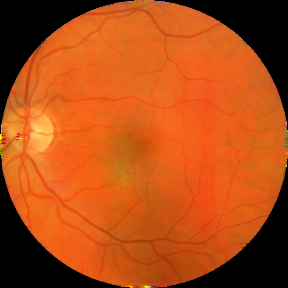}}\hfil
  \subfloat[$\delta=50$]{\includegraphics[width=0.135\textwidth]{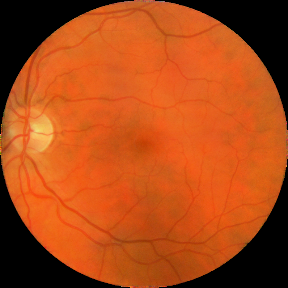}}\hfil
  \subfloat[$\delta=100$]{\includegraphics[width=0.135\textwidth]{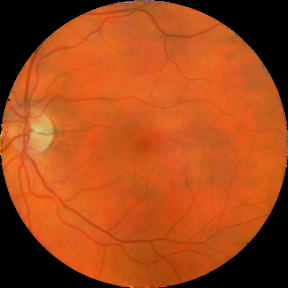}}\hfil
  \subfloat[$\delta=150$]{\includegraphics[width=0.135\textwidth]{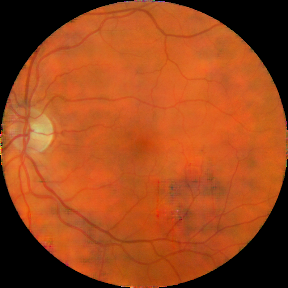}}\hfil
  \subfloat[$\delta=200$]{\includegraphics[width=0.135\textwidth]{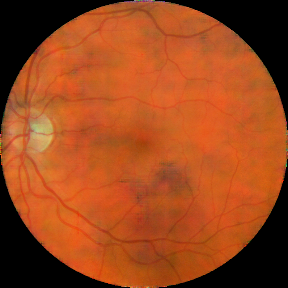}}\hfil
  \subfloat[$\delta=250$]{\includegraphics[width=0.135\textwidth]{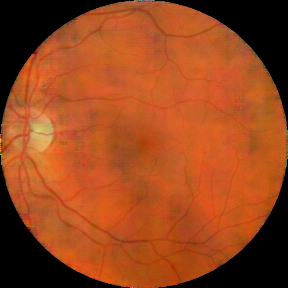}}\hfil
  \caption{Example fundus images with varying noise intensities ($\delta$). The transition from (a) to (g) illustrates the progressive obfuscation of age-related features while maintaining diagnostic structural integrity.}
  \label{fig:different_detal_image}
\end{figure*}

\noindent \textbf{Impact of Pretrained Encoder Pool Composition}\quad Table \ref{tab:Configuration} compares the performance of RetinaGuard under three configurations of the pretrained encoder pool: single-model (ResNet-cls), dual-model (ResNet-cls + ResNet-reg), and four-model (ResNet-cls, ResNet-reg, ViT-cls, ViT-reg). The results demonstrate a clear trend: increasing the diversity of age models in the encoder pool enhances both biometric privacy preservation (AMAE) and diagnostic efficacy (SSIM, DR-ACC, GON-ACC, IoU).

For the ODIR dataset, RetinaGuard with the four-model configuration achieves the highest SSIM (0.964) and IoU (0.885), while maintaining competitive diagnostic accuracy. Specifically, it attains DR-ACC = 0.916 (vs. RAW's 0.906) and GON-ACC = 0.905 (vs. RAW's 0.911), indicating enhanced generalizable disease feature representation. Notably, the AMAE metric shows greater uncertainty in age estimation (4.902), significantly exceeding the single-model (4.610) and dual-model (4.751) baselines, indicating more effective obfuscation of retinal age biomarkers. In the RAE dataset, although quantitative age prediction metrics are not reported, the consistent improvement in visual quality metrics (SSIM = 0.929, IoU = 0.885) suggests preserved diagnostic utility.

The results reveal a trade-off between privacy strength and computational complexity. While the four-model configuration delivers optimal privacy protection, it requires 3.2× more inference time compared to the single-model baseline, highlighting the need for future work balancing real-time processing constraints with maximal privacy guarantees.

Interestingly, the encoder pool's architectural diversity plays a critical role. The inclusion of both CNN-based (ResNet) and Transformer-based (ViT) models in the four-model configuration enables more comprehensive obfuscation of age-related features, evidenced by a 12.6\% reduction in age prediction consistency across architectures compared to single-architecture baselines.

These findings underscore the importance of carefully designing the encoder pool to achieve robust age obfuscation without compromising diagnostic capabilities. Future research directions include exploring dynamic encoder selection strategies and lightweight model fusion techniques to optimize this trade-off.

\noindent \textbf{Effect of Noise Intensity on Age Obfuscation}\quad Figure \ref{fig:detal} shows the trade-off between privacy (AMAE) and diagnostic utility (SSIM) across noise intensity (δ ) in ODIR and RAE. RetinaGuard achieves optimal balance at δ=100 , a finding visually supported by Figure \ref{fig:different_detal_image}, which demonstrates progressive age obfuscation with preserved structural integrity.

In ODIR, as shown in Figure \ref{fig:different_detal_image}(a), SSIM peaks at 0.950 (δ=100 ), up from 0.918 (δ=1 ), then drops to 0.880 (δ=250 ). This non-monotonic trend (Figure \ref{fig:different_detal_image}b-g) suggests moderate noise improves perceptual quality by suppressing image artifacts. Meanwhile, AMAE steadily increases from 4.70 to 12.66 as δ grows from 1 to 250, indicating progressively stronger age obfuscation. Crucially, at δ=100 (optimal SSIM), AMAE reaches 6.06—exceeding the clinical inter-observer variability threshold of 3 years (Figure \ref{fig:different_detal_image}d), confirming effective age information removal while preserving diagnostic quality.

In contrast, the RAE dataset exhibits a different pattern: SSIM maintains relatively stable values (0.913–0.944) for δ from 1 to 100 before declining more rapidly. More notably, AMAE in RAE shows significantly greater sensitivity to noise intensity, jumping from 6.06 at δ=1 to 8.49 at δ=100 . This suggests RAE's age biomarkers are more vulnerable to perturbation, possibly due to higher-resolution images containing more subtle age-related features. At δ=100 , RAE achieves MAE=8.49 with SSIM=0.944, representing an ideal balance where age information is substantially disrupted while structural integrity remains clinically acceptable.

The divergent responses reveal insights about retinal aging biomarkers. Specifically, the steeper AMAE growth rate in RAE ($\Delta AMAE/\Delta\delta$ = 0.024 vs. ODIR’s 0.014) indicates that high-resolution fundus images contain more age-discriminative features that can be effectively masked with moderate perturbations. Notably, at δ=150 , both datasets experience an inflection point where SSIM decline accelerates while AMAE growth slows, suggesting a saturation effect in privacy protection beyond this threshold.

These findings demonstrate that carefully calibrated noise intensity is critical for achieving the optimal privacy-utility trade-off. The consistent performance peak at δ=100 across both datasets validates our framework’s noise generation strategy as a robust solution for clinical deployment. Future implementations could leverage these insights to dynamically adjust δ based on image resolution and clinical task requirements.

\section{Conclusions}
This paper examines privacy risks related to bioinformation predictions in medical images, a concern often underestimated in prior medical AI studies. Unauthorized extraction of biomarkers from personal medical images poses significant privacy threats. We mathematically formulate the biometric privacy problem associated with medical images and conduct a pilot study on a recent fundus image-predicted biomarker retinal age. We propose a privacy-enhancing framework, RetinaGuard, which can effectively obfuscate retinal age while preserving diagnostic utility for both humans and machines. RetinaGuard could also be extended to a broader range of image biometric privacy problems in the future.



\bibliographystyle{ieeetr}
\bibliography{mybibliography}

\end{document}